# Defective Edge Detection Using Cascaded Ensemble Canny Operator

**Anjali Nambiyar [1]**
Research Department of Computer Science,
Bishop Heber College (Auto),
Affiliated to Bharathidasan University
Tiruchirappalli, India
anjalin.cs.res@bhc.edu.in

**Rajkumar Kannan [2]**
Research Department of Computer Science,
Bishop Heber College (Auto),
Affiliated to Bharathidasan University
Tiruchirappalli, India
rajkumar@bhc.edu.in

## ABSTRACT

Edge detection has been one of the most difficult challenges in computer vision because of the difficulty in identifying the borders and edges from the real-world images including objects of varying kinds and sizes. Methods based on ensemble learning, which use a combination of backbones and attention modules, outperformed more conventional approaches, such as Sobel and Canny edge detection. Nevertheless, these algorithms are still challenged when faced with complicated scene photos. In addition, the identified edges utilizing the current methods are not refined and often include incorrect edges. In this work, we used a Cascaded Ensemble Canny operator to solve these problems and detect the object edges. The most difficult Fresh and Rotten and Berkeley datasets are used to test the suggested approach in Python. In terms of performance metrics and output picture quality, the acquired results outperform the specified edge detection networks

## 1 Introduction

One of the most difficult tasks for computer vision applications is identifying the important edge in natural images. It's also the process of giving names to the borders that separate similar areas in a picture. Computing vision tasks like as object identification [1], picture segmentation [2], and picture reconstruction using spectra-aware screening mechanism (SASM) [3] may make use of this discovery. Furthermore, all of the suggested approaches are sensitive to changes in size, object form, and picture area intensity when evaluating detection quality. Researchers came up with several deep learning architectures [4–6] to tackle the above problems. The proposed work aims to detect the edges of the object using a unique Cascaded Ensemble Canny (CEC) operator. This article subsequent parts are organized as follows. In Section II, we provide relevant prior works; in Section III, we present the implemented proposed approach of edge detection; and in Section IV, we present the results and discussion. The article concluded in Section V.

## 2 Related Work

The identification of edges, borders, or contours is a challenging task in computer vision. Deep learning or statistical techniques are often used to tackle it. The suggested approach sought to extract the edge at each network block by operating on scale representations [7]. A Recursive Encoder-Decoder Network (REDN) for Edge Detection approach takes a different tack than previous techniques that used unified networks built from blocks of convolutional layers [8]. The Pixel Difference Network (PiDiNet) for efficient edge detection in a deep learning architecture. This approach makes has four separate blocks formed by pooling layers [9]. The edge detection map is the end result of all the processes. A model that is suggested for edge detection of VGG-16 used for feature extraction [10]. The final edge map is the result of integrating all the outputs. By preprocessing the pictures before feeding them into the model, the authors of [11] suggested a deep learning approach to Lightweight Dense Convolutional neural network for edge detection (LDC) [12].

## 3 Proposed Work

### 3.1 Edge Detection Process

The classification of edges is based on strength, distinguishing strong edges from weak ones. This reduces the likelihood of false positives by retaining only the most significant edges. And the edge enhancement via gradient computation utilizing the CEC operators summarize our approach in Figure 1 edge identification.

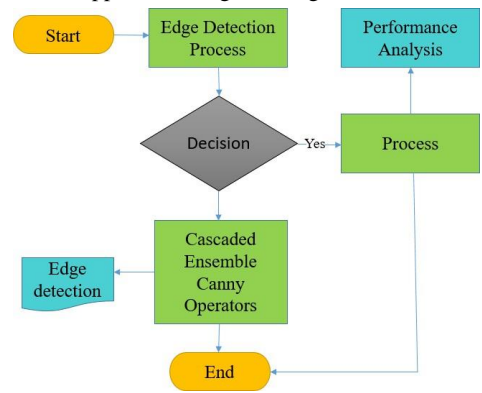

**Figure 1: Schematic representation of the suggested network**

Applied high and low thresholds, pixels with gradient magnitudes above this value are classified as strong edges. Finally, an image with strong, weak, and non-edge pixels is identified.

### 3.2 Cascaded Ensemble Canny Resolution Operator

CEC operator is on Quaternion-Based Canny Edge Detection filtering is an advanced edge detection method that extends the classical canny edge detector to work with quaternion-valued images. This approach is useful for processing color images or multidimensional signals. The following quaternion multiplication:

$$q_h = [\begin{array}{cc} 0 & R^* \end{array}]f_q(p,q)[\begin{array}{cc} R^* & 0 & R \end{array}]$$
$$q_v = [\begin{array}{c} 0 \\ R^* \end{array}]f_q(p,q)[\begin{array}{c} 0 \\ R \end{array}] \quad (1)$$

Where R is defined as the absolute value of the original quaternion-encoded picture $(p, q)$, where μ is a unit vector in three dimensions represented by a pure unit quaternion and $R = e^{\mu\theta}$.

Our focus here is on, $\mu = \frac{i+j+k}{\sqrt{3}}$ and $\theta = \frac{\pi}{2}$

Both directions of our approach make use of the modulus, which is defined as:

$$q_1(p,q) = \sqrt{q^2 f_{q_{hi}}(p,q) + q^2 f_{q_{hj}}(p,q) + q^2 f_{q_{hk}}(p,q)}$$
$$q_2(p,q) = \sqrt{q^2 f_{q_{vi}}(p,q) + q^2 f_{q_{vj}}(p,q) + q^2 f_{q_{vk}}(p,q)} \quad (2)$$

Thus the modulus grows in direct proportion to:

$$M(p,q) = \sqrt{q^2 f_{q_1}(p,q) + q^2 f_{q_2}(p,q)} \quad (3)$$

In order to extract the most essential information from a picture, the gradient is used to improve the image's edge points. Therefore, the average observation may be expressed as a matrix $g_{b \times n} = 1/b$, which summarises the picture. The formula $\mu = \frac{1}{b} g$ gives the standard deviation of an observation.

$$\mathbf{X} = \frac{1}{b}\sum_{b=1}^{b}(g_u - \mu)(g_u - \mu)^R \quad (4)$$

The test data vector and the mean-adjusted input data vector, denoted as Φ = R- μ.

$$\Phi_t = (R - \mu)^R = \Phi Y Y^R \quad (5)$$

An appropriate thresholding to find the locations with the sharpest contrast by selecting the most marked pixels from the initial operation. A two-dimensional first derivative is the vector that represents the gradient.

$$[f(p,q)] = \begin{bmatrix} G_x \\ G_y \end{bmatrix} = \begin{bmatrix} \frac{\partial f}{\partial x} \\ \frac{\partial f}{\partial y} \end{bmatrix} \quad (6)$$

The threshold gradient modulus of the edges is calculated as:

$$(p,q) = \tan^{-1}\left(\frac{G_y}{G_x}\right) \quad (7)$$

Finally the edge was calculated and enhanced.

## 4 Performance Analysis

### 4.1 Description of Datasets

The outcomes of our method's testing on the Berkley and Bangkit-JKT2-D datasets are detailed below in the references. Our technique relies on the Berkeley segmentation dataset, which contains all the pictures and ground facts. The results of object detection and in fruit defects edges simulation as a whole are shown in Figure 2.

**Figure 2: Simulated output**

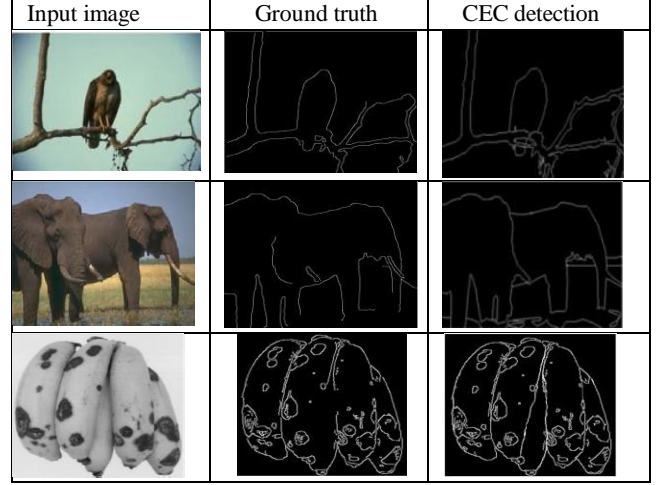

### 4.2 Comparative Analysis

Applying some of the performance indicators allows one to demonstrate the efficacy of the proposed technique, is done using specifications such as a) Accuracy and b) Specificity in Table 2.

**Table 1 Comparative Performance Analysis**

| Methods | Accuracy(%) | Specificity(%) |
|---|---|---|
| Sobel [1] | 87.8 | 95.8 |
| ERRNet [2] | 86.8 | 97.8 |
| SASM [3] | 94 | 94 |
| PiDiNet [9] | 78 | 86 |
| LCD [12] | 88 | 93 |
| CEC | 99 | 98 |

According to the results shown in figure 2 and table 2, the proposed approach achieves a high level of accuracy (99.8%) when compared to other systems already in use.

## 5 Conclusion

A novel approach to edge detection was proposed in the research. We were able to decrease the processing time and complexity of two-dimensional approaches in image processing by converting the edge detection issue into an image processing problem. In order to distinguish between edge and non-edge pixels, the ensemble method is also used. With each iteration, it aligns the leftover vector with atoms from the selected dictionary. When compared to other edge detectors on the market, the findings demonstrate the suggested one's efficient performance with a high range of accuracy (99.8%). This technique can identify weak edge pixels on defective fruit images using CEC operator. Additionally, we want to demonstrate encouraging outcomes in more imaging for the purpose of early illness diagnosis in the near future.